\documentclass[sigconf]{acmart}
\usepackage{subcaption}
\usepackage{graphicx}  

\AtBeginDocument{%
  \providecommand\BibTeX{{%
    \normalfont B\kern-0.5em{\scshape i\kern-0.25em b}\kern-0.8em\TeX}}}

\setcopyright{iw3c2w3g}
\copyrightyear{2021}
\acmYear{2021}
\acmDOI{10.1145/3465416.3483305}

\acmConference[EAAMO '21]{EAAMO '21: Equity and Access in Algorithms, Mechanisms, and Optimization}{October 5--9, 2021}{--, NY, USA}
\acmISBN{978-1-4503-8553-4/21/10}

\begin{document}

\title{A Framework for Understanding Sources of Harm throughout the Machine Learning Life Cycle}

\author{Harini Suresh}\email{hsuresh@mit.edu}
\author{John Guttag}\email{guttag@mit.edu}

\renewcommand{\shortauthors}{Suresh and Guttag}


\begin{abstract}
As machine learning (ML) increasingly affects people and society, awareness of its potential unwanted consequences has also grown. To anticipate, prevent, and mitigate undesirable downstream consequences, it is critical that we understand when and how harm might be introduced throughout the ML life cycle. 
In this paper, we provide a framework that identifies seven distinct potential sources of downstream harm in machine learning, spanning data collection, development, and deployment. In doing so, we aim to facilitate more productive and precise communication around these issues, as well as more direct, application-grounded ways to mitigate them. 


\end{abstract}

\maketitle

\section{Introduction} 
Machine learning (ML) is increasingly used to make decisions that affect people's lives. Typically, ML algorithms operate by learning models from existing data and generalizing them to unseen data.  As a result, problems can arise during the data collection, model development, and deployment processes that can lead to different harmful downstream consequences.  In recent years, we have seen such examples in diverse contexts such as facial analysis (e.g., where publicly available algorithms performed significantly worse on dark-skinned women \citep{buolamwini2018gender}) and pre-trial risk assessment of defendants in the criminal justice system (e.g., where a deployed algorithm was more likely to incorrectly predict black defendants as being high-risk \cite{angwin2016machine}).


A common refrain is that undesirable behaviors of ML systems happen when ``data is biased.'' Indeed, a recent comment by a prominent ML researcher\footnote{https://twitter.com/ylecun/status/1274782757907030016?s=20} to this end set off a heated debate\,---\, not necessarily because the statement ``data is biased'' is \textit{false}, but because it treats data as a static artifact divorced from the process that produced it. This process is long and complex, grounded in historical context and driven by human choices and norms. Understanding the implications of each stage in the data generation process can reveal more direct and meaningful ways to prevent or address harmful downstream consequences that overly broad terms like ``biased data'' can mask. 

Moreover, it is important to acknowledge that not all problems should be blamed on the data. The ML pipeline involves a series of choices and practices, from model definition to user interfaces used upon deployment. Each stage involves decisions that can lead to undesirable effects. For an ML practitioner working on a new system, it is not straightforward to identify if and how problems might arise. Even once identified, it is not clear what the appropriate application- and data-specific mitigations might be, or how they might generalize over factors such as time and geography.  

Consider the following simplified scenario: a medical researcher wants to build a model to help detect whether someone is having a heart attack. She trains the model on medical records from a subset of prior patients at a hospital, along with labels indicating if and when they suffered a heart attack. She observes that the system has a higher false negative rate for women (it is more likely to miss cases of heart attacks in women), and hypothesizes that the model was not able to effectively learn the signs of heart attacks in women because of a lack of such examples. She seeks out additional data representing women who experienced heart attacks to augment the dataset, re-trains the model, and observes that the performance for female patients improves.  Meanwhile, a co-worker hiring new lab technicians tries to build an algorithm for predicting the suitability of a candidate from their resume along with human-assigned ratings.  He notices that women are much less likely to be predicted as suitable candidates than men. Like his colleague, he tries to collect many more samples of women to add to the dataset, but is disappointed to see that the model's behavior does not change. Why did this happen? The \textit{sources} of the disparate performance were different. In the first case, it arose because of a lack of data on women, and introducing more data was helpful. In the second case, using human assessment of quality as a label to estimate true qualification allowed the model to discriminate by gender, and collecting more labelled data from the same distribution did not help.  


This paper provides a framework and vocabulary for understanding distinct \textit{sources} of downstream harm from ML systems.  In practice, we imagine this framework being used in different ways by a variety of stakeholders, including those who build, evaluate, use, or are affected by ML systems. We demonstrate how issues arise in distinct stages of the ML life cycle, and provide corresponding terminology that avoids overly broad and/or overloaded terms.  Doing so can facilitate a more methodical analysis of the risks of a particular system, and can inform appropriate mitigations that stem from an application-grounded understanding of the data generation and development processes. Thinking prospectively, the framework can also help practitioners anticipate these issues and design more thoughtful and contextual methods for data collection, development, evaluation, and/or deployment. Beyond those involved in model development, an understanding of how and why issues arise throughout the ML life cycle can provide a valuable guide for external stakeholders, such as regulators or affected populations, to question or probe a system.


Throughout the paper, we refer to the concept of ``harm'' or ``negative consequences'' caused by ML systems. \citet{barocas2017problem} provide a useful framework for thinking about how these consequences actually manifest, splitting them into \textit{allocative} harms (when opportunities or resources are withheld from certain people or groups) and \textit{representational} harms (when certain people or groups are stigmatized or stereotyped). For example, algorithms that determine whether someone is offered a loan or a job \citep{de2019bias,raghavan2020mitigating} risk inflicting allocative harm. This is typically the type of harm that we think and hear about, because it can be measured and is more commonly recognized as harmful. However, even if they do not directly withhold resources or opportunities, systems can still cause representational harm; e.g., language models that encode and replicate stereotypes.  

Section 2 follows with a brief overview of the ML pipeline that will be useful background information as we refer to different parts of this process. Section 3 details each source of harm in more depth with examples. In Section 4, we provide a more rigorous presentation of our framework for formalizing and mitigating the issues we describe. Finally, Section 5 is a brief conclusion. 



\begin{figure*}[t] 
	\centering
	\begin{subfigure}{\textwidth} 
		\includegraphics[width=\textwidth]{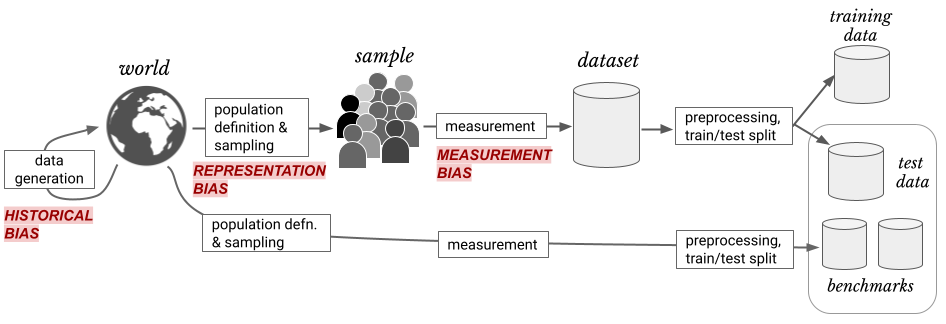}
		\vspace{-20pt}
		\caption{Data Generation} 
	\end{subfigure}
	\vspace{1em} 
	\begin{subfigure}{\textwidth} 
		\includegraphics[width=\textwidth]{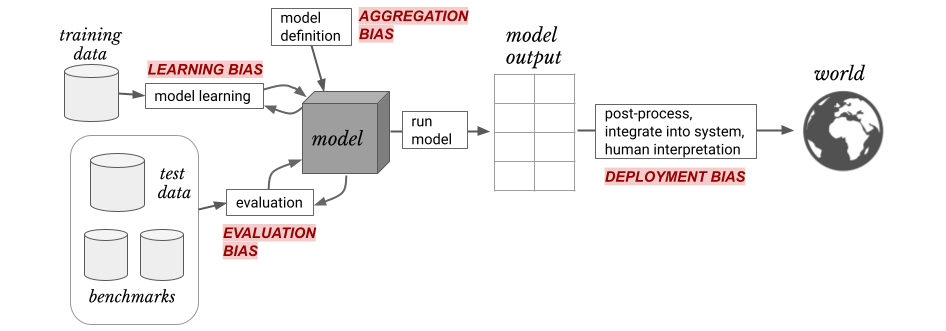}
		\vspace{-26pt}
		\caption{Model Building and Implementation} 
	\end{subfigure}
	\caption{(a) The data generation process begins with data collection. This process involves defining a target population and sampling from it, as well as identifying and measuring features and labels. This dataset is split into training and test sets.  Data is also collected (perhaps by a different process) into benchmark datasets. (b) A model is defined, and optimized on the training data. Test and benchmark data is used to evaluate it, and the final model is then integrated into a real-world context.  This process is naturally cyclic, and decisions influenced by models affect the state of the world that exists the next time data is collected or decisions are applied. In red, we indicate where in this pipeline different sources of downstream harm might arise.} \label{fig:pipeline}
\end{figure*}

\section{Machine Learning Overview} 
Machine learning is a type of statistical inference that learns, from existing data, a function that can be generalized to new, unseen data. ML algorithms are all around us: making personalized Netflix or YouTube recommendations, powering Siri's stilted conversation, providing live transcriptions on our video calls, auto-tagging the people in our photos, deciding whether we are offered job interviews, or approving (or not) tests at the doctor's office. In each of these examples, an ML algorithm has found patterns in a (usually massive) dataset, and is applying that knowledge to make a prediction about new data points (which might be photos, medical records, resumes, etc.).  

In this section, we will briefly describe the typical life cycle of an ML system.  We will describe each step generally, as well as how it might occur in a running hypothetical example: a machine learning-based loan-approval system. In the running example, we describe each step as it typically happens (not necessarily as it ideally \textit{should}). In the next section, we analyze the implications of each step and problems that may be introduced. 
Figure \ref{fig:pipeline} depicts these steps. Later, in Section 4, we provide a more rigorous formalization of these steps.

\subsection*{Data Collection} Before any analysis or learning happens, data must first be collected.  Compiling a dataset involves identifying a \textit{target population} (of people or things), as well as defining and measuring \textit{features} and \textit{labels} from it.  Typically, it is not feasible to include the entire target population, and instead, features and labels are sampled from a subset of it (here, we refer to this subset as the \textit{development sample}). Often, ML practitioners use existing datasets rather than going through the data collection process. 

\textbf{Example.} For the loan approval system, a team in charge of data collection could choose the target population to be people who live in the state in which the system will be used, people who have previously applied for loans, people with credit cards, etc.  The particular sample that ends up in the dataset will be a subset of this target population and will depend upon the sampling method (e.g., sourcing information from public records or surveying people). 
There is also the question of what to actually measure or collect about these people: perhaps things like their debt history, the number of credit cards they have, their income, their occupation, etc.  Some of these things will be chosen to serve as labels: for example, information about whether the person received and/or paid back a loan in the past.  

\subsection*{Data Preparation} Depending on the data modality and task, different types of preprocessing may be applied to the dataset before using it. Datasets are usually split into \textit{training data} used during model development, and \textit{test data} used during model evaluation.  Part of the training data may be used as \textit{validation data}. 

\textbf{Example.} For the loan approval system, preprocessing might involve dealing with missing data (e.g., imputing missing credit history values via interpolation), simplifying the feature space (e.g., grouping occupations in broader categories like ``physician'' rather than encoding detailed specialities), or normalizing continuous measurements (e.g., scaling income so it lies on a 0-to-1 scale). If a resulting dataset included 1000 examples (e.g., data collected from 1000 people), 600 examples might be used for training, 100 as a validation set during training, and 300 for post-development testing.  

\subsection*{Model Development} Models are then built using the training data (not including the held-out validation data).  Typically, models are trained to optimize a specified \textit{objective}, such as minimizing the mean squared error between its predictions and the actual labels. A number of different model types, hyperparameters, and optimization methods may be tested out at this point; usually these different configurations are compared based on their performance on the validation data, and the best one chosen.  

\textbf{Example.} The team developing the loan approval model would  first need to instantiate a particular model (e.g., a dense, feed-forward neural network) and define an objective function (e.g., minimizing the cross-entropy loss between the model's predictions and the label defined in the training data). Then, in the optimization process, the model will try to learn a function that goes from the inputs (e.g., income, occupation, etc.) to the output (e.g., whether the person paid back a previous loan). They might also train a number of different models (e.g., with varying architectures or training procedures) and choose the one that performs best on the validation set. 

\subsection*{Model Evaluation}
After the final model is chosen, the performance of the model on the test data is reported.  The test data is not used before this step, to ensure that the model's performance is a true representation of how it performs on unseen data. Aside from the test data, other available datasets --- also called \textit{benchmark datasets} --- may be used to demonstrate model robustness or to enable comparison to other existing methods. The particular \textit{performance metric(s)} used during evaluation are chosen based on the task and data characteristics.

\textbf{Example.} Here, the model developed in the previous step would be evaluated by its performance on the test set. There might be several performance metrics to consider\,---\,for example, applicants might be concerned with false negatives (i.e., being denied a loan when they actually are deserving), while lenders might care more about false positives (i.e., recommending loans to people who don't pay them back). In addition, the model might be evaluated on existing datasets used for similar tasks (e.g., the dataset from the U.S. Small Business Association described in \citet{li2019loan}).

\subsection*{Model Postprocessing}
Once a model has been trained, there are various post-processing steps that may needed. For example, if the output of a model performing binary classification is a probability, but the desired output to display to users is a categorical answer, there remains a choice of what threshold(s) to use to round the probability to a hard classification. 

\textbf{Example.} The resulting model for predicting loan approval likely outputs a continuous score between 0 and 1.  The team might choose to transform this score into discrete buckets (e.g., low-risk of defaulting, unsure, high-risk of defaulting) or a binary recommendation (e.g., should/should not receive a loan).  
 
\subsection*{Model Deployment}
There are many steps that arise in deploying a model to a real-world setting.  For example, the model may need to be changed based on requirements for explainability or apparent consistency of results, or there may need to be built-in mechanisms to integrate real-time feedback. Importantly, there is no guarantee that the population a model sees as input after it is deployed (here, we will refer to this as the \textit{use population}) looks the same as the population in the development sample. 

\textbf{Example.} In order to deploy the loan approval system, the team will likely need to develop a user interface that displays the result and the recommended action. They might need to develop different visualizations of the model's reasoning and results for lenders, applicants, regulatory agencies, or other relevant stakeholders. And they may need to incorporate mechanisms for applicants to seek recourse if they believe the model recommendation was inaccurate or discriminatory.

\section{Seven Sources of Harm in ML}

In this section, we will go into more depth on potential sources of harm. There are several possible organizational principles for creating such a taxonomy. For example, \citet{ntoutsi2020bias} distinguish issues caused by data generation, data collection, or institutional bias; and \citet{mehrabi2021survey} group types of bias based on how they interact with the data, the algorithm, or the user.  Here, with the goal of focusing on \textit{sources} of harm, we choose to use the different stages in the ML life cycle for organizational structure; the sources of harm we describe roughly map to the processes described in Figure 1. Each subsection will detail where and how in the ML life cycle problems might arise, as well as a characteristic example. These categories are not mutually exclusive; however, identifying and characterizing each one as distinct makes them less confusing and easier to tackle.

We use the term ``bias'' to describe these problems primarily because of precedence, acknowledging that it is a heavily overloaded term that is used to describe a range of issues across different fields. Here, the biases we describe refer to distinct sources of harm in an ML system, and can be thought of as breaking down vague terms like ``algorithmic bias'' or ``data bias'' into more useful and granular concepts. Types of bias conceptualized in other works might map onto our framework depending on where in the ML life cycle they manifest. For example, ``cognitive bias,'' in crowd annotators of a dataset would fall under the umbrella of our ``measurement bias'' (Section 3.3), because it describes an issue that arises during the process of measuring labels in a dataset. Similarly, \citet{friedman1996bias}'s ``preexisting bias'' might map to our ``historical bias'' (Section 3.1) when it describes existing societal stereotypes that are reflected in datasets.

\subsection{Historical Bias}
Historical bias arises even if data is perfectly measured and sampled, if the world \textit{as it is} or \textit{was} leads to a model that produces harmful outcomes. Such a system, even if it reflects the world accurately, can still inflict harm on a population. Considerations of historical bias often involve evaluating the representational harm (such as reinforcing a stereotype) to a particular group.  

\subsubsection{Example: Word Embeddings}
Word embeddings are learned vector representations of words that encode semantic meaning, and are widely used for natural language processing (NLP) applications. Recent research has shown that word embeddings, which are learned from large corpora of text (e.g., Google news, web pages, Wikipedia), reflect human biases. One such study  \cite{garg2018word} demonstrates that word embeddings reflect real-world biases about women and ethnic minorities, and that an embedding model trained on data from a particular decade reflects the biases of that time.  For example, gendered occupation words like ``nurse'' or ``engineer'' are highly associated with words that represent women or men, respectively. A range of NLP applications (e.g., chatbots, machine translation, speech recognition) are built using these types of word embeddings, and as a result can encode and reinforce harmful stereotypes. 


\subsection{Representation Bias}
Representation bias occurs when the development sample under-represents some part of the population, and subsequently fails to generalize well for a subset of the use population. Representation bias can arise in several ways: 
\begin{enumerate} 
\item \textbf{When defining the target population, if it does not reflect the use population.} Data that is representative of Boston, for example, may not be representative if used to analyze the population of Indianapolis. Similarly, data representative of Boston 30 years ago will likely not reflect today’s population.
\item \textbf{When defining the target population, if contains under-represented groups.} Say the target population for a particular medical dataset is defined to be adults aged 18-40. There are minority groups within this population: for example, people who are pregnant may make up only 5\% of the target population. Even we sample perfectly, and even if the use population is the same (adults 18-40), the model will likely be less robust for those 5\% of pregnant people because it has less data to learn from.  
\item \textbf{When sampling from the target population, if the sampling method is limited or uneven.} For example, the target population for modeling an infectious disease might be all adults, but medical data may be available only for the sample of people who were considered serious enough to bring in for further screening. As a result, the development sample will represent a skewed subset of the target population. In statistics, this is typically referred to as \textit{sampling bias}.
\end{enumerate}

\subsubsection{Example: Geographic Diversity in Image Datasets}
ImageNet is a widely-used image dataset consisting of 1.2 million labeled images \citep{imagenet_cvpr09}. ImageNet is intended to be used widely (i.e., its target population is ``all natural images''). However, ImageNet does not evenly sample from this target population; instead, approximately 45\% of the images in ImageNet were taken in the United States, and the majority of the remaining images are from North America or Western Europe.  Only 1\% and 2.1\% of the images come from China and India, respectively.  As a result, \citet{imagenetdiversity} show that the performance of a classifier trained on ImageNet is significantly worse at classifying images containing certain objects or people (such as ``bridegroom'') when the images come from under-sampled countries such as Pakistan or India.

\subsection{Measurement Bias}
Measurement bias occurs when choosing, collecting, or computing features and labels to use in a prediction problem. Typically, a feature or label is a \textit{proxy} (a concrete measurement) chosen to approximate some \textit{construct} (an idea or concept) that is not directly encoded or observable. For example, ``creditworthiness'' is an abstract construct that is often operationalized with a measureable proxy like a credit score. Proxies become problematic when they are poor reflections or the target construct and/or are generated differently across groups, which can when:
\begin{enumerate}
    \item \textbf{The proxy is an oversimplification of a more complex construct.} Consider the prediction problem of deciding whether a student will be successful (e.g., in a college admissions context).  Fully capturing the outcome of ``successful student'' in terms of a single measurable attribute is impossible because of its complexity.  In cases such as these, algorithm designers may resort to a single available label such as ``GPA'' \citep{kleinberg2018algorithmic}, which ignores different indicators of success present in different parts of the population. 
    \item \textbf{The method of measurement varies across groups.} For example, consider factory workers at several different locations who are monitored to count the number of errors that occur (i.e., observed number of errors is being used as a proxy for work quality). If one location is monitored much more stringently or frequently, there will be more errors observed for that group. This can also lead to a feedback loop wherein the group is subject to further monitoring because of the apparent higher rate of mistakes \citep{barocas2016big,pmlr-v81-ensign18a}.
    \item \textbf{The accuracy of measurement varies across groups.}  For example, in medical applications, ``\textit{diagnosed} with condition X'' is often used as a proxy for ``has condition X.'' However, structural discrimination can lead to systematically higher rates of misdiagnosis or underdiagnosis in certain groups \cite{mossey2011defining,phelan2015impact,hoffmann2001girl}. For example, there are both gender and racial disparities in diagnoses for conditions involving pain assessment \citep{calderone1990influence,hoffman2016racial}. 
\end{enumerate}

\subsubsection{Example: Risk Assessments in the Criminal Justice System} 
Risk assessments have been deployed at several points within criminal justice settings \cite{henry2019risk}. For example, risk assessments such as Northpointe's COMPAS predict the likelihood that a defendant will re-offend, and may be used by judges or parole officers to make decisions around pre-trial release \citep{angwin2016machine}. The data for models like these often include proxy variables such as ``arrest'' to measure ``crime'' or some underlying notion of ``riskiness.''  Because minority communities are more highly policed, this proxy is \textit{differentially mismeasured}\,---\,there is a different mapping from crime to arrest for people from these communities.  Many of the other features used in COMPAS (e.g., ``rearrest'' to measure ``recidivism'' \citep{dressel2018accuracy}) were also differentially measured proxies. The resulting model had a significantly higher false positive rate for black defendants versus white defendants (i.e., it was more likely to predict that black defendants were at a high-risk of reoffending when they actually were not). 

\subsection{Aggregation Bias}
Aggregation bias arises when a one-size-fits-all model is used for data in which there are underlying groups or types of examples that should be considered differently. Underlying aggregation bias is an assumption that the mapping from inputs to labels is consistent across subsets of the data. In reality, this is often not the case.  A particular dataset might represent people or groups with different backgrounds, cultures or norms, and a given variable can mean something quite different across them. Aggregation bias can lead to a model that is not optimal for any group, or a model that is fit to the dominant population (e.g., if there is also representation bias). 



\subsubsection{Example: Social Media Analysis.}
\citet{patton2020contextual} describe analyzing Twitter posts of gang-involved youth in Chicago. By hiring domain experts from the community to interpret and annotate tweets, they were able to identify shortcomings of more general, non-context-specific NLP tools. For example, certain emojis or hashtags convey particular meanings that a nonspecific model trained on all Twitter data would miss. In other cases, words or phrases that might convey aggression elsewhere are actually lyrics from a local rapper \cite{frey2020artificial}. Ignoring this group-specific context in favor of a single, more general model built for all social media data would likely lead to harmful misclassifications of the tweets from this population.


\subsection{Learning Bias}
Learning bias arises when modeling choices amplify performance disparities across different examples in the data \cite{hooker2021moving}. For example, an important modeling choice is the objective function that an ML algorithm learns to optimize during training. Typically, these functions encode some measure of accuracy on the task (e.g., cross-entropy loss for classification problems or mean squared error for regression problems). However, issues can arise when prioritizing one objective (e.g., overall accuracy) damages another (e.g., disparate impact) \cite{kleinberg_et_al:LIPIcs:2017:8156}. For example, minimizing cross-entropy loss when building a classifier might inadvertently lead to a model with more false positives than might be desirable in many contexts.   


\subsubsection{Example: Optimizing for Privacy or Compactness.} Recent work has explored training models that maintain \textit{differential privacy} (i.e., preventing them from inadvertently revealing excessive identifying information about the training examples during use). However, \citet{bagdasaryan2019differential} show that differentially private training, while improving privacy, reduces the influence of underrepresented data on the model, and subsequently leads to a model with worse performance on that data (as compared to a model without differentially private training). Similarly, \citet{hooker2020bias} show how prioritizing compact models (e.g., with methods such as \textit{pruning}) can amplify performance disparities on data with underrepresented attributes. This happens because, given limited capacity, the model learns to preserve information about the most frequent features.

\subsection{Evaluation Bias}
Evaluation bias occurs when the benchmark data used for a particular task does not represent the use population. A model is optimized on its training data, but its quality is often measured on benchmarks (e.g., UCI datasets\footnote{https://archive.ics.uci.edu/ml/datasets.php}, Faces in the Wild\footnote{http://vis-www.cs.umass.edu/lfw/}, ImageNet\footnote{http://www.image-net.org/}). This issue operates at a broader scale than other sources of bias: a misrepresentative benchmark encourages the development and deployment of models that perform well only on the subset of the data represented by the benchmark data.

Evaluation bias ultimately arises because of a desire to quantitatively compare models against each other.  Applying different models to a set of external datasets attempts to serve this purpose, but is often extended to make general statements about how good a model is.  Such generalizations are often not statistically valid \citep{salzberg1997comparing}, and can lead to overfitting to a particular benchmark. This is especially problematic if the benchmark suffers from historical, representation or measurement bias.  

Evaluation bias can also be exacerbated by the choice of metrics that are used to report performance. For example, aggregate measures can hide subgroup underperformance, but these singular measures are often used because they make it more straightforward to compare models and make a judgment about which one is ``better.'' Just looking at one type of metric (e.g., accuracy) can also hide disparities in other types of errors (e.g., false negative rate). 


\subsubsection{Example: Commercial Facial Analysis Tools} \citet{buolamwini2018gender} point out the drastically worse performance of commercial facial analysis algorithms (performing tasks such as gender- or smiling- detection) on images of dark-skinned women. Images of dark-skinned women comprise only 7.4\% and 4.4\% of common benchmark datasets Adience and IJB-A, and thus benchmarking on them failed to discover and penalize underperformance on this part of the population. Since this study, other algorithms have been benchmarked on more balanced face datasets, changing the development process itself to encourage models that perform well across groups \citep{ryu2018inclusivefacenet}. 

\subsection{Deployment Bias}
Deployment bias arises when there is a mismatch between the problem a model is intended to solve and the way in which it is actually used. This often occurs when a system is built and evaluated as if it were fully autonomous, while in reality, it operates in a complicated sociotechnical system moderated by institutional structures and human decision-makers (\citet{selbst2019fairness} refers to this as the ``framing trap''). In some cases, for example, systems produce results that must first be interpreted by human decision-makers. Despite good performance in isolation, they may end up causing harmful consequences because of phenomena such as automation or confirmation bias.  

\subsubsection{Example: Risk Assessment Tools in Practice}
Algorithmic risk assessment tools in the criminal justice context (also described in Section 3.3.1) are models intended to predict a person's likelihood of committing a future crime. In practice, however, these tools may be used in ``off-label'' ways, such as to help determine the length of a sentence. \citet{collins2018punishing} describes the harmful consequences of risk assessment tools for actuarial sentencing, including justifying increased incarceration on the basis on personal characteristics.  \citet{stevenson2018assessing} builds on this idea, and through an in-depth study of the deployment of risk assessment tools in Kentucky, demonstrates how evaluating the system in isolation created unrealistic notions of its benefits and consequences.  

\subsection{Identifying Sources of Harm}
Knowledge of a model's context and intended use should inform identifying and addressing sources of harm. Recognizing historical bias, for example, requires a retrospective understanding of how structural oppression and discrimination has manifested in a particular domain over time.  Issues that arise in image recognition are frequently related to representation or evaluation bias since many large publicly-available image datasets and benchmarks are collected via web scraping, and thus do not equally represent different groups, objects, or geographies. When features or labels represent human decisions (e.g., diagnoses in the medical context, human-assigned ratings in the hiring context), they typically serve as proxies for some underlying, unmeasurable concepts, and can introduce measurement bias. Identifying aggregation bias usually requires some understanding of meaningful underlying groups in the data and reason to think they have different conditional distributions with respect to the prediction label. Medical applications, for example, often risk aggregation bias because patients of different sexes with similar underlying conditions may present and progress in different ways. Deployment bias is often a concern when systems are used as decision aids for people, since the human intermediary may act on predictions in ways that are typically not modeled in the system. 

\section{Formalization and Mitigations}
\subsection{Formalizing the framework}
\begin{figure}[h]
{\includegraphics[width=\linewidth]{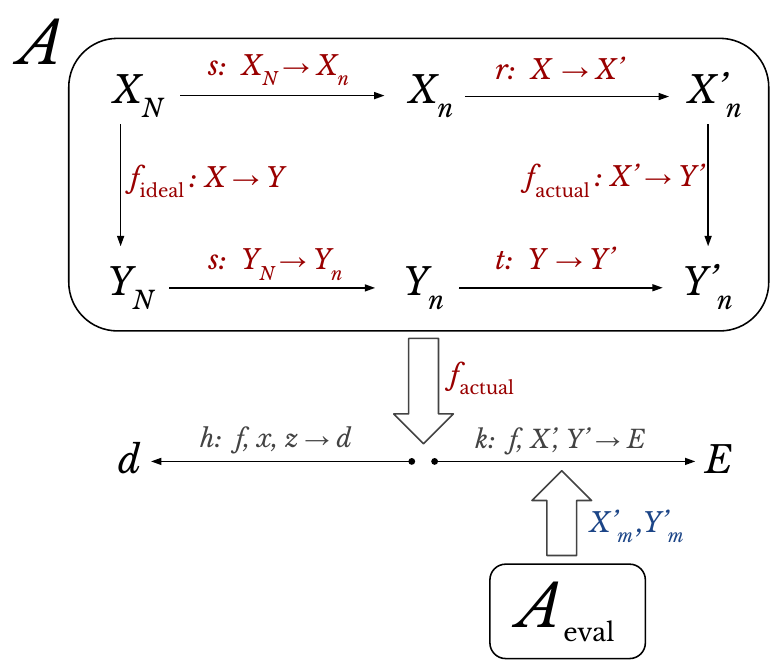}}
{\caption{A data generation and ML pipeline viewed as a series of mapping functions. The upper part of the diagram deals with data collection and model building, while the bottom half describes the evaluation and deployment process. See the text for a detailed walk-through.}\label{fig:diagram}}
\end{figure}

We now take a step towards formalizing some of the notions introduced in the previous sections. We do this by abstracting the ML pipeline to a series of data transformations. This formalization provides a context we then use to discuss targeted mitigations for specific sources of bias. 

Consider the data transformations for a dataset as depicted in Figure \ref{fig:diagram}. This data transformation sequence can be abstracted into a general process $A$. Let $X$ and $Y$ be the underlying feature and label constructs we wish to capture.  The subscript indicates the size of the populations, so $X_N$ indicates these constructs over the target population and $X_n$ indicates the smaller development sample, where $s: {X}_N \rightarrow {X}_n$ is the sampling function.  $X'$ and $Y'$ are the measured feature and label proxies that are chosen to build a model, where $r$ and $t$ are the projections from constructs to proxies, i.e., $X \rightarrow X'$ and $Y \rightarrow Y'$. The function $f_{\text{ideal}}: X \rightarrow Y$ is the target function---learned using the ideal constructs from the target population---but $f_{\text{actual}}: X' \rightarrow Y'$ is the actual function that is learned using proxies measured from the development sample.  Then, the function $k$ computes some evaluation metric(s) $E$ for $f_{\text{actual}}$ on data $X'_m, Y'_m$ (possibly generated by a different process, e.g., $A_{\text{eval}}$ in Figure \ref{fig:diagram}).  Finally, given the learned function $f_{\text{actual}}$, a new input example $x$, and any external, environmental information $z$, a function $h$ governs the real-world decision $d$ that will be made (e.g., a human decision-maker taking a model's prediction and making a final decision). 

\subsection{Designing Mitigations}
There is a growing body of work on ``fairness-aware algorithms'' that modify some part of the modeling pipeline to satisfy particular notions of ``fairness.'' Interested readers are referred to \citet{narayanan2018fat} for a detailed overview of different fairness definitions typically found in this literature, and \citet{benchmarks} for a comparison of several of these techniques on a number of different datasets. \citet{finocchiaro2021bridging} further discuss potential issues and mitigation mechanisms in the context of a range of application domains. Here, our aim is to understand and motivate mitigation techniques in terms of their ability to target different \textit{sources} of harm. In doing so, we can get a better understanding when and why different approaches might help, and what hidden assumptions they make. Understanding where intervention is necessary and how feasible it is can also inform discussions around when harm can be mitigated versus when it is better not to deploy a system at all. 

We do not include a table or checklist of mitigations for different problems here, to avoid implying that there is a comprehensive or generalizable set of solutions. Instead, we intend this framework to provide a useful organizational structure for thinking through potential problems, understanding if and what mitigation techniques are appropriate, and/or motivating new ones.

As an example, measurement bias is related to how features and labels are generated (i.e., how $r$ and $t$ are instantiated). Historical bias is defined by inherent problems with the distribution of $X$ and/or $Y$ across the entire population.  Therefore, solutions that try to adjust $s$ by collecting more data (that then undergoes the same transformation to $X'$) will likely be ineffective for either of these issues.  However, it may be possible to combat historical bias by designing $s$ to systematically over- or under-sample $X$ and $Y$, leading to a development sample with a different distribution that does not reflect the same undesirable historical biases. In the case of measurement bias, changing $r$ and $t$ through more thoughtful, context-aware measurement or annotation processes (e.g., as in \citet{patton2020contextual}) may work. 

In contrast, representation bias stems either from the target population definition ($X_N$, $Y_N$) or the sampling function ($s$). In this case, methods that adjust $r$ or $t$ (e.g., choosing different features or labels) or $f$ (e.g., changing the objective function) may be misguided.  Importantly, solutions that \textit{do} address representation bias by adjusting $s$ implicitly assume that $r$ and $t$ are acceptable and that therefore, improving $s$ will mitigate the harm. 

Aggregation bias is a limitation on the learned function $f$ that stems from an assumption about the homogeneity of $p(Y' | X')$, and can result in an $f$ that is disproportionately worse for some group(s). Therefore, aggregation bias could be targeted through 1) parameterizing $f$ so that it better models the data complexities (e.g., coupled learning methods, such as multitask learning, that take into account group differences \citep{pmlr-v81-dwork18a,suresh2018multitask}), or 2) transforming the training data such that $f$ is now better suited to it (e.g., projecting data into a learned representation space where $p(Y' | X')$ is the same across groups \citep{zemel2013learning}). Note that methods that attempt to make predictions independently of group membership \citep{corbett_defining} likely will not address aggregation bias.  

Learning bias is an issue with the way $f$ is optimized, and mitigations should target the defined objective(s) and learning process \cite{hooker2021moving}.  In addition, some sources of harm are connected: e.g., learning bias can exacerbate performance disparities on under-represented groups, so changing $s$ to more equally represent different groups/examples could also help prevent it. 

Evaluation bias is an issue with $E$, which is a measure of the quality of the learned function, $f$. Tracing the inputs to $E$, we can see that addressing it could involve 1) redefining $k$ (the function that computes evaluation metrics) and/or 2) adjusting the data $X'$ and $Y'$ on which metrics are computed. We might improve $k$ through computing and reporting a broader range of metrics on more granular subsets of the data (e.g. as in Gender Shades \citep{buolamwini2018gender}). The best groups and metrics to use are often application-specific, requiring intersectional analysis and privacy considerations; they should be chosen with domain specialists and affected populations that understand the usage and consequences of the model. In a predictive policing application, for example, law enforcement may prioritize a low false negative rate (not missing any high-risk people) while affected communities may value a low false positive rate (not being mistakenly classified as high-risk). See \citet{modelcards} for a more in-depth discussion. Issues with evaluation data $X'_m$ and $Y'_m$ stem from an problems within the data generation process in $A_{\text{eval}}$, e.g., an unrepresentative sampling function $s_{\text{eval}}$. Improving $s_{\text{eval}}$ could involve targeted data augmentation to populate parts of the data distribution that are underrepresented \citep{chawla2002smote, chen2018why}. In other cases, it may be better to develop entirely new benchmarks that are more representative and better suited to the task at hand \cite{denton2020bringing, karkkainen2021fairface,atwood2020inclusive}.

Deployment bias arises when $h$ introduces unexpected behavior affecting the final decision $d$. Dealing with deployment bias is challenging since the function $h$ is usually determined by complex real-world institutions or human decision-makers. Mitigating deployment bias might involve systems that help users balance their faith in model predictions with other information and judgements \cite{jacovi2021formalizing}. This might be facilitated by choosing an $f$ that is human-interpretable, and/or by developing intuitive interfaces that help users understand model uncertainty and how predictions should be used.

\section{Conclusion}
This paper provides a framework for understanding the sources of downstream harm caused by ML systems.  We do so in a way that we hope will facilitate productive communication around these issues; we envision future work being able to state upfront which particular type of bias they are addressing, making it immediately clear what problem they are trying to solve and what assumptions they are making about the data and domain. 

By framing sources of downstream harm through the data generation, model building, evaluation, and deployment processes, we encourage application-appropriate solutions rather than relying on broad notions of what is fair.  Fairness is not one-size-fits-all; knowledge of an application and engagement with its stakeholders should inform the identification of these sources. 

Finally, we illustrate that there are important choices being made throughout the broader data generation and ML pipeline that extend far beyond just model training. In practice, ML is an iterative process with a long and complicated feedback loop. We highlight problems that manifest through this loop, from historical context to the process of benchmarking models to their final integration into real-world processes.

\bibliographystyle{ACM-Reference-Format}
\bibliography{main.bib}

\end{document}